\documentclass[conference]{IEEEtran}
\IEEEoverridecommandlockouts
\usepackage{cite}
\usepackage{amsmath,amssymb,amsfonts}
\usepackage{algorithmic}
\usepackage{graphicx}
\usepackage{textcomp}
\usepackage{xcolor}
\def\BibTeX{{\rm B\kern-.05em{\sc i\kern-.025em b}\kern-.08em
    T\kern-.1667em\lower.7ex\hbox{E}\kern-.125emX}}
    \usepackage[left=0.68in,right=0.68in,top=0.75in,bottom=1.1in]{geometry}
\begin{document}

\title{Depth Monocular Estimation with Attention-based Encoder-Decoder Network from Single Image 
%\\
}

\author{\IEEEauthorblockN{1\textsuperscript{st} Xin Zhang}
\IEEEauthorblockA{%\textit{Electrical Engineering} \\
\textit{University of South Carolina}\\
Columbia, United States \\
xz8@email.sc.edu}
\and
\IEEEauthorblockN{2\textsuperscript{nd} Rabab Abdelfattah}
\IEEEauthorblockA{%\textit{Electrical Engineering} \\
\textit{University of South Carolina}\\
Columbia, United States \\
rabab@email.sc.edu}
\and
\IEEEauthorblockN{3\textsuperscript{rd} Yuqi Song}
\IEEEauthorblockA{%\textit{Computer Science Engineering} \\
\textit{University of South Carolina}\\
Columbia, United States \\
yuqis@email.sc.edu}
\and
 \IEEEauthorblockN{4\textsuperscript{th} Samuel A. Dauchert}
 \IEEEauthorblockA{%\textit{Electrical Engineering} \\
 \textit{University of South Carolina}\\
 Columbia, United States \\
 dauchert@email.sc.edu}
\and
\IEEEauthorblockN{5\textsuperscript{th} Xiaofeng Wang}
\IEEEauthorblockA{%\textit{Electrical Engineering} \\
\textit{University of South Carolina}\\
Columbia, United States \\
wangxi@cec.sc.edu}
}
%\thanks{The authors gratefully acknowledge the partial financial support of the
%National Science Foundation (1830512 and 2018966) .}
\maketitle

%to do list:
%弄清楚数据集
%找一份可以运行的代码

%不生成全部的depth图，size自定义

%全size的深度图可以用很多现有的benchmark
%随距离，随遮挡程度

%attention based，对比的是distance estimation
%随距离，随遮挡程度

%depth + distance,前者用于别的场景，后者适用于av？

\begin{abstract}

Depth information is the foundation of perception, essential for autonomous driving, robotics, and other 
source-constrained applications. Promptly obtaining accurate and efficient depth information allows for 
a rapid response in dynamic environments. Sensor-based methods using LIDAR and RADAR obtain high precision 
at the cost of high power consumption, price, and volume. While due to advances in deep learning, 
vision-based approaches have recently received much attention and can overcome these drawbacks. In this
work, we explore an extreme scenario in vision-based settings: estimate a depth map from one monocular 
image severely plagued by grid artifacts and blurry edges.
To address this scenario,
We first design a convolutional attention mechanism block (CAMB) which consists of channel attention and spatial attention sequentially and insert these CAMBs into skip connections. As a result, our novel approach can find the focus of current image with minimal overhead and avoid losses of depth features.
Next, by combining the depth value, the gradients of X axis, Y axis and diagonal directions, and the structural similarity index measure (SSIM), we propose our novel loss function. Moreover, we utilize 
pixel blocks to accelerate the computation of the loss function. Finally, we show, through comprehensive experiments on two large-scale image datasets, i.e. KITTI and NYU-V2, that our method outperforms several representative baselines.

\end{abstract}


\section{Introduction}
%自动驾驶中感知距离重要，其他获取距离的设备的缺点。基于图像值得研究。
%双目常见，但是也有问题。因此单目值得研究。
%单目研究的方法分类，我们的方法属于xx。为什么其他类不使用？
%本文主要利用xx，解决了xx问题。尤其是对xx的考虑，这在之前类似工作都没有。
%本文的主要贡献为：
%文章结构如下。

Perception is one of the key technologies in many areas, such as autonomous driving, virtual reality, and robotics~\cite{van2018autonomous}, which helps to detect, understand, and interpret the surrounding environments, including dynamic and static obstacles. 
The performance of perception usually relies on the accuracy of depth information estimation~\cite{el2019survey}. 
For example, 
autonomous driving requires to estimate the inter-vehicle distance and warn potential rear-end collisions~\cite{zhe2020inter}, 
robotic arms cannot grasp the target without accurate depth information~\cite{hwang2020applying}, and so on.

%depth map的子问题 distance，可以加一个图来说明

There exist many strategies to infer depth information.  In general, these strategies can be classified into two categories: sensor-based methods and image-based methods~\cite{ming2021deep,zhe2020inter}. 
Sensor-based strategies, such as utilizing like LIDAR, RGB-D camera, and other active sensors~\cite{daniels2017forward}, are able to collect depth information accurately.
However, this type of methods usually places heavy burdens on manpower and computation~\cite{zhang2020monocular}.  %,mueller2019real
In addition, there could be strict conditions when applying these methods.  For instance, LIDAR estimates depth accurately only at sparse locations~\cite{badki2021binary} and RGB-D camera suffers from its limited measurement range and outdoor sunlight sensitivity~\cite{zhao2020monocular}.
Alternatively, image-based methods can overcome these issues and be applied in a wide range of applications~\cite{sabnis2011single, liu2014discrete}.
The conventional image-based depth estimation methods heavily rely on multi-view geometry~\cite{schonberger2016structure, basha2012structure, javidnia2017accurate}, such as stereo images~\cite{scharstein2007learning, scharstein2002taxonomy} and consecutive frames. Nevertheless, it introduces issues such as calibration drift over time~\cite{badki2021binary,el2019survey} as well as high demands on computational resources and memory~\cite{khan2020deep}. 
Therefore, using a monocular camera becomes an alternative low-cost, efficient, and attractive solution with light maintenance requirements for autonomous driving, robotics, and other resource-constrained applications~\cite{sabnis2011single}.

\begin{figure}[!tbp]
\centerline{\includegraphics[width=0.5\textwidth]{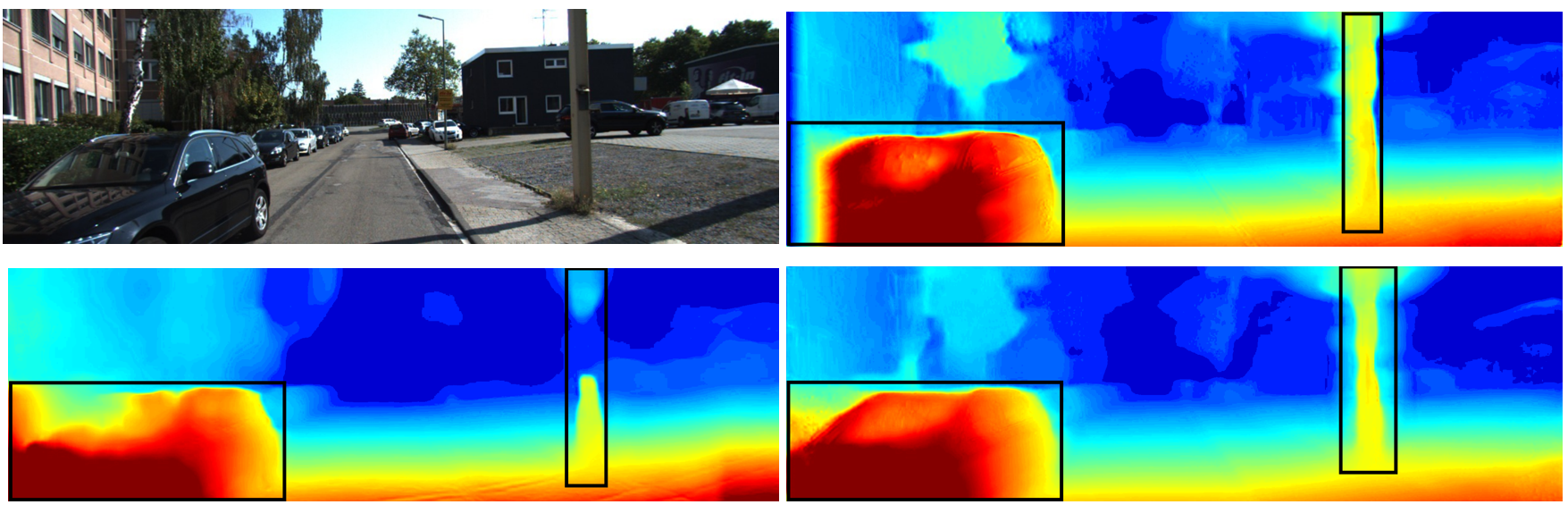}}
\caption{Generated depth map of different methods. Upper left is an image in the KITTI dataset.
The upper right and lower left images are generated by~\cite{godard2017unsupervised} and~\cite{yin2018geonet} respectively. It can be seen that the objects in these two images (cars, poles, framed by black boxes) are obviously incomplete and blurred. Lower right is ours. }
\label{fig1}
\end{figure}

This paper studies the extreme case in monocular depth estimation, which is to estimate the depth map from one image.  This could be an ill-posed problem as there is an ambiguity in the scale of the depth~\cite{khan2020deep}.
Owing to the release of publicly available datasets and the advancement of Convolutional Neural Networks (CNNs), Eigen et al.~\cite{eigen2014depth} first prove that the scale information can be learned by properly designing the network structure~\cite{eigen2014depth}.
After this, there has been a lot of work along this direction~\cite{yin2019enforcing, xu2017multi}.
Despite their success, there are still some critical issues to be addressed: 
\begin{itemize}

    \item Many methods do not consider the contextual information and treat all pixels equally. It may result in the grid artifacts problem~\cite{chen2019rethinking} and the edges in depth maps may be distorted or blurry~\cite{hu2019revisiting}, as shown in Figure~\ref{fig1}.
    %缺图片举例解释
    \item Depth estimation is often deeply integrated with industrial applications, which require real-time operation with limited computational resources.
    In order to achieve higher accuracy, however, deeper networks and complex mechanisms are developed with more parameters~\cite{ming2021deep}.
    The conflict between real-time requirements and expensive computational overhead should be mitigated urgently. 
    \item For traditional CNN architecture, such as fully connected network (FCN), after multiple layers of information processing, the depth features could be severely lost, which may lead to low accuracy and cannot meet the requirements in practice~\cite{ming2021deep}.
\end{itemize}

%This is an ill-posed problem as there is an ambiguity in the scale of the depth~\cite{khan2020deep, eigen2014depth}.
%With the release of publicly available datasets and the advancement of Convolutional Neural Networks (CNNs), it has been prove that the scale information can be learned by properly designing the network structure~\cite{eigen2014depth, yin2019enforcing, xu2017multi}.
%However, for traditional CNN architecture, such as fully connected network (FCN), after multiple layers of information processing, the depth features are severely lost, which lead to low accuracy and cannot meet the requirements of practical applications~\cite{ming2021deep}.
%In addition, many methods do not consider constraints in real-world scenarios, and the contextual information is ignored, which may result in grid artifacts problem~\cite{chen2019rethinking},
%for example, %说一下图片中的问
%What's more, depth estimation is closely integrated with industrial applications, which have high requirements for the real-time performance within limited computing resources. 
%However, in order to pursue higher accuracy, researchers tend to construct deeper networks and complex mechanisms with more parameters~\cite{ming2021deep}.
%The conflict between real-time requirements and expensive computational overhead needs to be mitigated urgently. 

To alleviate these issues, this paper presents an new approach for depth monocular estimation from a single image.  
The main contributions are summarized as follows:
\begin{itemize}

%我们提出了一个基于注意力的ende模型，能够有效的根据单张图片生成相应的深度图。
%
%大量的实验，在两个数据集上证明了我们方法的有效性。
\item We propose an encoder-decoder attention based network to effectively generate corresponding depth map from a single image and avoid grid artifacts with least possible overhead. To leverage the contextual information and find focuses of images, we design a convolutional attention mechanism block (CAMB) by combining channel attention and spatial attention sequentially and insert these CBAMs into the skip connections. Different from many of the previous methods, our attention module is light-weight and therefore more suitable for resource-constrained applications.

\item We design a novel loss function by combining the depth value, the gradients of three dimensions (i.e. X-axis, Y-axis and diagonal direction) and structural similarity index measure (SSIM). In addition, we introduce pixel blocks, instead of single pixel, to save computational resources when calculating the loss.

\item We conduct comprehensive experiments on two large-scale datasets, i.e. KITTI and NYU-V2.  It is shown that our approach outperforms several representative baseline methods, which verify the effectiveness of our approach.

    %\item We propose an encoder-decoder attention based network to effectively generate corresponding depth map from a single image. 
    %In order to preserve the scale information during learning, we add skip connections between the corresponding layers of the encoder and decoder. 
    %Then we insert convolutional block attention module (CBAM), which is a light-weight attention module, into skip connections to help our network to find the focus and avoid grid artifacts with least possible overhead. 
    %\item We design a novel loss function by combining the depth value and the gradients of three dimensions and structural similarity index measure (SSIM). In addition, we use pixel blocks, instead of single pixel, to save computational resources when calculating the loss. 
    %\item We conduct comprehensive experiments on two large-scale datasets, i.e. KITTI and NYU-V2.  It is shown that our approach outperforms several representative baseline methods, which verify the effectiveness of our approach.
\end{itemize}

%lightweight

%很多方法都是通过完整的深度图来还原距离，但是注意力不集中，多了额外的计算。
%我们的方法不仅能够完整深度图，还能够集中注意力。

The remainder of this paper is organized as follows. 
After a brief review of related work in Section~\ref{subsec:rw}, we present the monocular depth estimation problem in Section~\ref{subsec:fl}. 
Section~\ref{subsec:pm} proposes our new approach. We evaluate the qualitative and quantitative performance on KITTI and NYU-v2 in Section~\ref{subsec:ex}.
Finally, conclusions are drawn in Section~\ref{subsec:con}.
\begin{figure*}[t]
  \centering
  \includegraphics[width=0.93\linewidth]{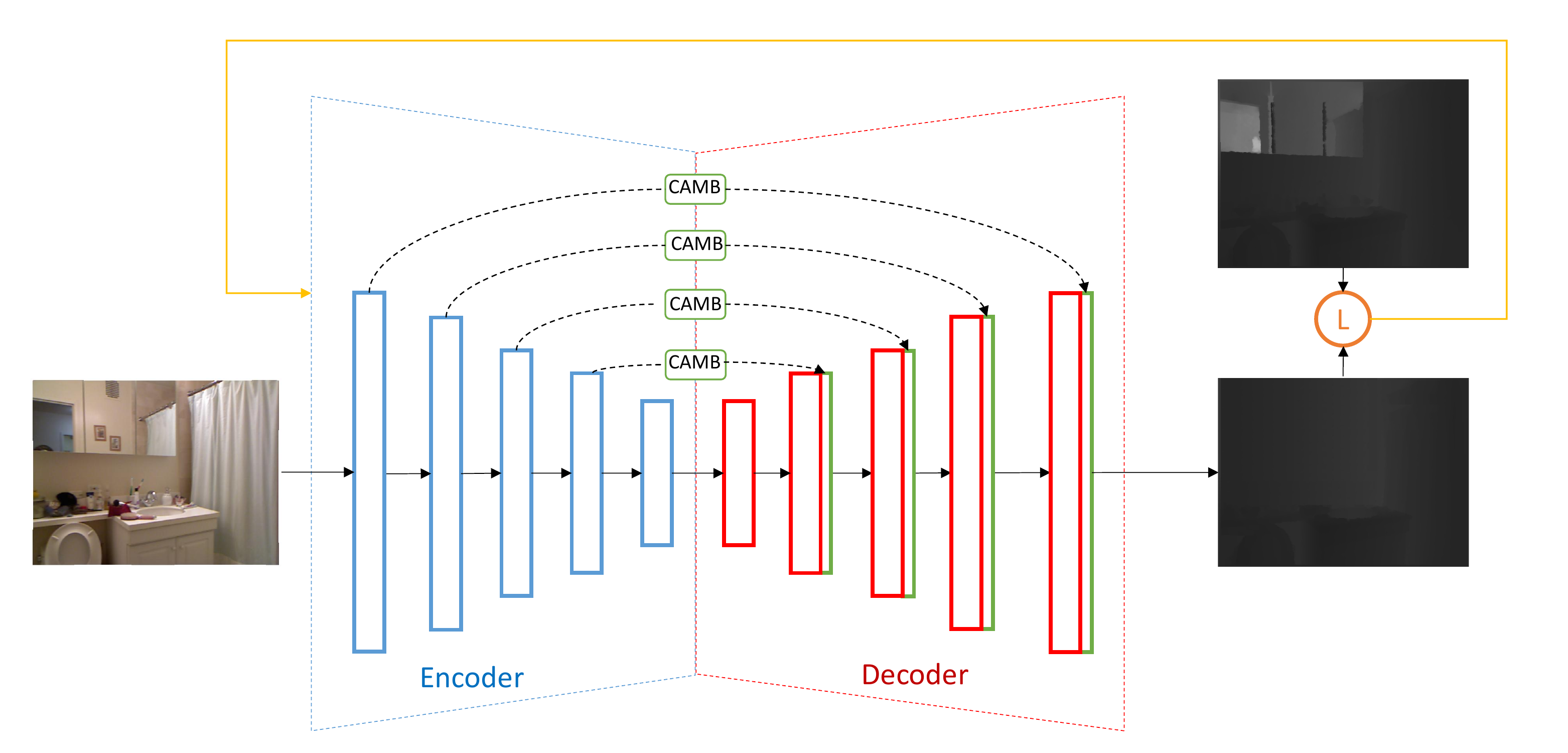}
  \caption{The architecture of our model. CAMB and L stands for our attention module and loss function as Equation~\ref{overall} respectively. It can be seen that after inserting CAMB to skip connection (dotted lines), our model, our model captures both normal (red blocks) and attention features (green blocks). The yellow line on the top stands for the backpropagation. Due to the lightweight property of CAMB, our approach is able to add attention information with least possible overhead, while the skip connections ensure that scale information is preserved even in later stage of learning.}
  \label{fig:network}
\end{figure*}

\section{Related Work}
\label{subsec:rw}
Recently, numerous methods have been proposed for image-based depth estimation. 
We can roughly divide these methods into two categories: geometry-based and monocular. 

\subsection{Geometry-based methods}
Recovering 3D structures based on geometric constraints is an optional method to estimate depth information.
This kind of methods relies on consecutive frames taken by one camera or stereo matching based on binocular camera.
For the former, structure from motion (SfM)~\cite{ullman1979interpretation} is a representative method by matching features of different frames and estimating the camera motion, but the performance heavily relies on the quality of image sequences~\cite{zhao2020monocular}. 
To alleviate this problem, a variety of sfM strategies has been proposed to deal with uncalibrated or unordered images~\cite{schonberger2016structure}.
For example, incremental sfM approaches~\cite{snavely2006photo, wu2013towards} add on one image at a time to grow the reconstruction, global methods~\cite{wilson2014robust} consider the entire view graph at the same time, and hierarchical methods~\cite{gherardi2010improving} divides the images into multiple clusters, reconstructs each cluster separately and merges partial models into a complete model. However, they still suffer from monocular scale ambiguity and high computational complexity~\cite{zhao2020monocular}.
As for the latter, it calculates the disparity maps~\cite{hamzah2016literature} of images through a cost function, and its bottleneck is the accuracy of matching the pixels of different images~\cite{chang2018real}. Different from sfM, the scale information is included in depth estimation since the cameras is calibrated in advance in this case~\cite{lazaros2008review}. 
However, in addition to the high consumption of computing and memory, calibration drift is also an issue~\cite{badki2021binary}. 

\subsection{Monocular methods from Single Image}
%由此，我们发现上述两类方法都存在共同的缺点：因此，我们使用基于单目的深度估计方法。
Since there is only one single image need to be calculated, depth estimation from one image can effectively reduce the computational complexity and memory overhead~\cite{badki2021binary}. 
Numerous methods have been proposed for estimating depth information from one image in recent years. Herein, we briefly review the relevant studies.

This problem was firstly studied by Eigen et al.~\cite{eigen2014depth}.
They regard this problem as a regression problem and propose a CNNs architecture which is composed of global coarse-scale network and local fine-scale network to generate depth maps.   
By taking advantage of the 3D geometric constraints, Yin et al.~\cite{yin2019enforcing} implement 'virtual norm' constraints~\cite{khan2020deep} and proposed a supervised framework to obtain a high-quality depth estimation. 
Praful et al.~\cite{hambarde2021uw} utilize UW-GAN to estimate depth information, their network includes two modules: the generator predicts depth maps, and the discriminator determines the quality of the maps. 
Fu et al.~\cite{fu2018deep} introduce a spacing-increasing discretization (SID) strategy to discretize depth and recast depth network learning as an ordinal regression problem to generate depth maps.
Xu et al.~\cite{xu2017multi} propose a conditional random field (CRF) based model for the multi-scale features to estiamte the fine-grained depth maps.
Although these fully connected network (FCN) based methods have achieved great success, there still exists some critical limitations,
such as inconsistent labeling, losing or smoothing object details and requiring extra memory for holding a large part of parameters~\cite{yasrab2017encoder, chen2021attention}.

\textbf{Encoder-Decoder. }
To alleviate these problems, encoder-decoder architecture with skip connections was proposed~\cite{sutskever2014sequence} and has made significant contributions in many vision related problems, which including estimating depth information.
In recent years, the use of such architecture have shown great success in improving overall performance of estimating depth maps.
%For removing the requirement that the inputs must be fixed dimensionality and reducing the dimension of images to facilitate depth estimation~\cite{sutskever2014sequence}, a huge number of works use encoder-decoder architecture to estimate depth maps and have shown great success in improving overall performance.
%which has shown great success in many vision related problems, such as image segmentation~\cite{ronneberger2015u} and optical flow estimation~\cite{dosovitskiy2015flownet} and image restoration~\cite{lehtinen2018noise2noise}. 
The encoder network usually consists of convolution and pooling layers to capture features, the decoder includes deconvolution layers to generate desired information, and the corresponding layers of encoder and decoder are concatenated with skip connections.
Alhashim et al.~\cite{alhashim2018high} employ a straightforward encoder-decoder architecture with no addition modifications and a Structural Similarity-based (SSIM) loss function to generate depth maps.  
Lee et al.~\cite{lee2019big} propose a monocular depth estimation method that uses new Local Planar Guidance Layers (LPGL) inserted into the decoding phase of the network. 
Fangchang et al.~\cite{ma2018sparse} use UpProj module~\cite{laina2016deeper} as upsampling layer in decoding layers and the encoder consist of a ResNet followed by a convolution layer.  

However, despite the above approaches make full use of the computational power of CNNs, they ignore the contextual information, in other words, they process all pixels equally which easily results in the grid artifacts problem~\cite{chen2019rethinking}.

%这一点补充一个图？？？？？？？？？？？？？

\textbf{Attention Mechanisms. }
To address this problem, we add attention mechanisms (AM) into our network.
AM was first applied to NLP problem, and then it has achieved great success in CNNs recently, which can help networks to focus on key objects and take advantage of contextual information~\cite{wang2016survey}.
From this perspective, our work is most closely related to~\cite{chen2019rethinking}, which propose an Attention-Based Context Aggregation Network (ACAN) which utilizes the deep residual architecture, dilated layer and self-attention to estimate depth.
It should be pointed that we keep the traditional encoder-decoder architecture with skip connections and attention module, and our AM is lightweight. 
Therefore, our model is more applicable to resource-constrained applications.
In addition, \cite{chen2019rethinking} adopt the sum of attention loss and ordinal loss added in a certain proportion as loss function, 
but we utilize Structural Similarity Index Measure (SSIM) and gradients between adjacent pixel blocks to propose a new loss function.

% 此外，与传统的注意力方法不同，我们的方法不需要xx，因此xxx

\section{Problem Representation}
\label{subsec:fl}
For an image $I\in\mathbb{R}^{H\times W\times C}$ with height $H$, width $W$ and channel $C$, the goal of depth estimation is to estimate the depth map~\cite{jeon2015accurate} $D\in\mathbb{R}^{H\times W}$, which has the same size with the original image but only one channel.%park2011high
However, due to the lack of global scale information, this task is inherently ambiguous and technically ill-posed~\cite{eigen2014depth}.
As the development of deep neural network and the improvement of publicly datasets' quality, we are able to address this issue by learning the scale information from training sets $\{(I_i, D_i)\}$~\cite{chen2021attention,alhashim2018high}. % li2018megadepth, zhou2020da4ad, 
Mathematically, we regard this problem as a regression problem which usually uses a standard loss function such as MSE~\cite{allen1971mean}.

\begin{figure*}[!ht]
  \centering
  \includegraphics[width=0.93\linewidth]{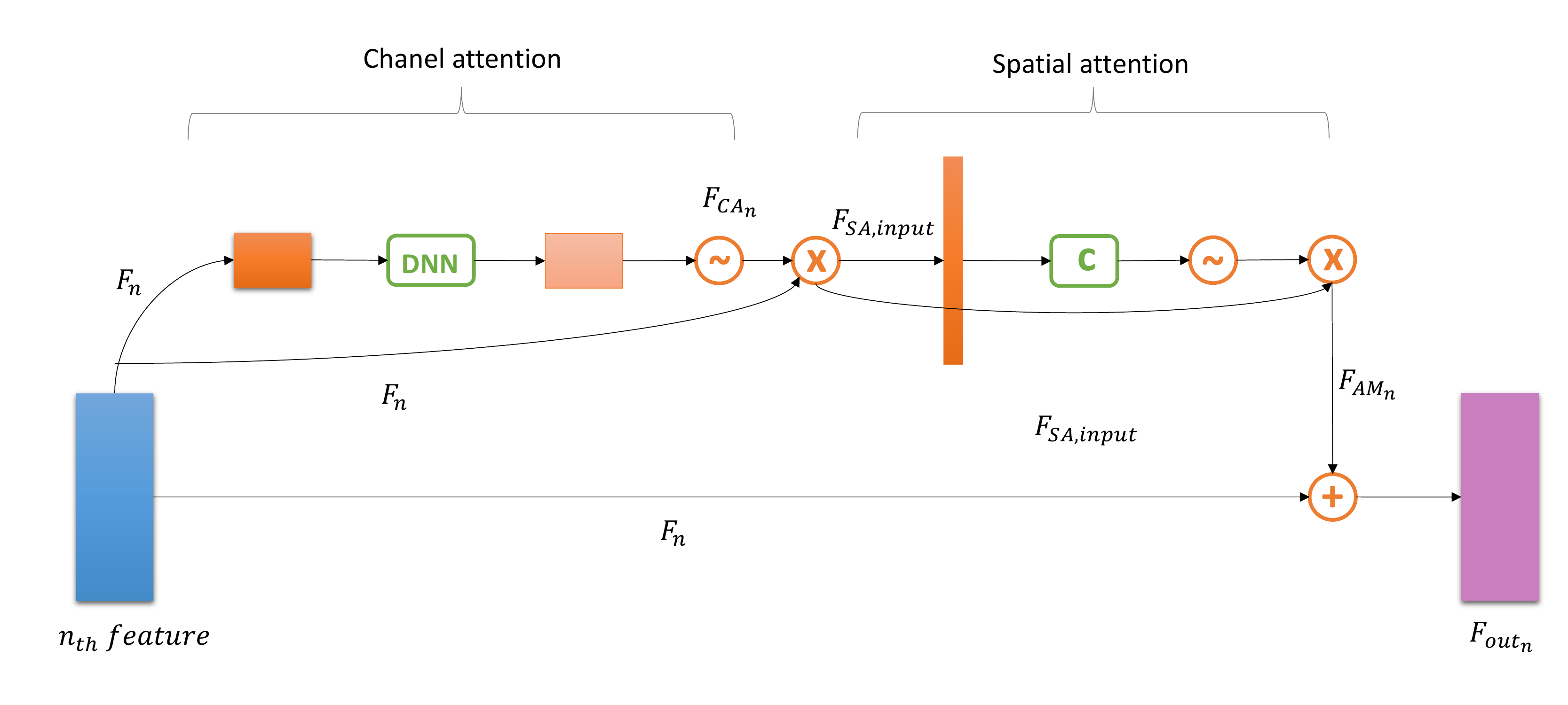}
  \caption{The pipeline of CAMB attention module. For the output feature $F_n$ of $n$th layer of encoder, we perform channel attention operation and spatial attention operation in sequence, and then add the calculation result $F_{AM_n}$ to $F_n$ to obtain the final feature $F_{out_n}$ that combines the normal feature with attention feature. }
  \label{fig:cbam}
\end{figure*}

\section{Proposed Method}
\label{subsec:pm}
We introduce the architecture of our attention-based encoder-decoder network and the design of loss function for the monocular depth estimation in this section.

\subsection{Network Architecture}
For an input RGB image $I_i$, our method is able to generate the corresponding depth map $D_i$ in an end-to-end fashion.
As shown in Figure~\ref{fig:network}, our network mainly consists of encoder and decoder. The corresponding layers of encoder and decoder are connected by skip connections with CAMBs. 
We use DesNet-169~\cite{huang2017densely} without the last classification layer as encoder, which extracts high-resolution features and downsamples the input image. Our encoder is pretrained on ImageNet dataset~\cite{deng2009imagenet}.
The decoder in our network contains a straightforward up-scaling scheme, which simply upsamples the output of the previous layer to the same size as the output of the corresponding encoder layer after CAMB, then concatenate these two output feature together and performs a convolution operation.

%具体而言，（介绍数据流向和处理的先后顺序）

\subsection{Attention Module}
%就像人类视觉一样，我们希望模型能够将注意力集中在重点
As mentioned before, the grid artifacts and blurry edges limit the performance of depth estimation.
To alleviate this problem, we design an attention module, which can help our network to pay different attention to different pixels, that is, our network is able to focus on the objects worthy of attention, and appropriately reduce the attention to the background, thereby reducing grids and fining edges. 
Considering practical application scenarios, we hope that our attention module can be light-weight so that it will not put extra burden on strained computing resources.

Therefore, based on CBAM~\cite{woo2018cbam} whose lightweight property has been well proved, 
we design a more lightweight and effective convolution attention mechanism block (CAMB) by and insert CAMBs into our model. 
Different from CBAM, our attention module utilize global power average pooling~\cite{estrach2014signal}, which is formulated as~\eqref{power} where $R$ stands for the current feature map and $p$ is the hyperparameter, and more simplified operations
\begin{equation}
    \tilde{a} = \sqrt[p]{\sum_{i\in R}{a_i^p}}. \label{power}
\end{equation}
It should be pointed out that when we set $p=1$ or $\infty$, equation~\eqref{power} actually represents sum pooling, which is proportional to average pooling, and max pooling.
%While in original CBAM, we have to calculate max pooling and average pooling in both channel attention and spatial attention separately.
Specifically, we do not directly pass the output feature $\textbf{F}_n$ of the encoder's $n$th layer to the corresponding layer of the decoder through skip connections. 
Instead, we first feed $\textbf{F}_n$ to CAMB, which has channel attention and spatial attention two sequential sub-modules.
The former performs global power average pooling operations on $\textbf{F}_n$, and passes the pooling results to a global shared three-layer fully connected DNN. Then execute the sigmoid function $s(\cdot)$ to get the channel feature map $\textbf{F}_{\text{CA}_n}$. 
%The former performs average pooling and max pooling operations on $\textbf{F}_n$, and passes the pooling results to a global shared multi-layer perception (MLP) respectively. Then using element-wise summation for these two outputs of MLP and execute the sigmoid function $s(\cdot)$ to get the channel feature map $\textbf{F}_{\text{CA}_n}$. 
%The latter uses the product of $\textbf{F}_{\text{CA}_n}$ and $\textbf{F}_n$ as input and performs average and max pooling along the channel axis. After concatenating these two pooling results, we execute a convolution operation $c(\cdot)$ to generate a feature map and pass it to the sigmoid function $s(\cdot)$ to get the final spatial attention map $\textbf{F}_{\text{SA}_n}$.
The latter uses the product of $\textbf{F}_{\text{CA}_n}$ and $\textbf{F}_n$ as input and performs power average pooling along the channel axis. Then we execute a convolution operation $c(\cdot)$ with kernel size of $7\times7$, which is experimentally determined, to generate a feature map and pass it to the sigmoid function $s(\cdot)$ to get the final spatial attention map $\textbf{F}_{\text{SA}_n}$.
In the end, we multiply $\textbf{F}_{\text{CA}_n}$ and $\textbf{F}_{\text{SA}_n}$ to get the final attention feature $\textbf{F}_{\text{AM}_n}$.

In order to add attention features on the basis of retaining the original features, we merge the final attention feature $\textbf{F}_{\text{AM}_n}$ of CAMB with $\textbf{F}_n$ using element-wise summation. Mathematically, we summarize the process of passing feature as follows:
\begin{align*}
    &\textbf{F}_{\text{CA}_n} = s({\rm DNN}({\rm pap}(\textbf{F}_n))  \in \mathbb{R}^{1\times 1\times C} \\
    &\textbf{F}_{\text{SA,input}_n} = \textbf{F}_{\text{CA}_n}\times \textbf{F}_n \in \mathbb{R}^{H\times W\times C} \\
    &\textbf{F}_{\text{SA}_n} = s(c({\rm pap}(\textbf{F}_{\text{SA,input}_n})) \in \mathbb{R}^{H\times W\times 1} \\
    &\textbf{F}_{\text{AM}_n} = \textbf{F}_{\text{SA}_n}\times \textbf{F}_{\text{SA,input}_n} \in \mathbb{R}^{H\times W\times C} \\
    &\textbf{F}_{\text{out}_n} = (\textbf{F}_{\text{AM}_n} + \textbf{F}_n) \in \mathbb{R}^{H\times W\times C}
\end{align*}
% \begin{align*}
%     &\textbf{F}_{\text{CA}_n} = s({\rm MLP}({\rm avg}(\textbf{F}_n)) + {\rm MLP}(\max(\textbf{F}_n)))  \in \mathbb{R}^{1\times 1\times C} \\
%     &\textbf{F}_{\text{SA,input}_n} = \textbf{F}_{\text{CA}_n}\times \textbf{F}_n \in \mathbb{R}^{H\times W\times C} \\
%     &\textbf{F}_{\text{SA}_n} = s(c([{\rm avg}(\textbf{F}_{\text{SA,input}_n});\max(\textbf{F}_{\text{SA,input}_n})])) \in \mathbb{R}^{H\times W\times 1} \\
%     &\textbf{F}_{\text{AM}_n} = \textbf{F}_{\text{SA}_n}\times \textbf{F}_{\text{SA,input}_n} \in \mathbb{R}^{H\times W\times C} \\
%     &\textbf{F}_{\text{out}_n} = \textbf{F}_{\text{AM}_n} + \textbf{F}_n \in \mathbb{R}^{H\times W\times C}
% \end{align*}
where $H$, $W$ and $C$ stands for height, weight and channel of images respectively, and $pap(\cdot)$ stands for the power average pooling operation.
The detail of this process is shown in Figure~\ref{fig:cbam}.

\subsection{Loss Function}
%MSE+polling difference (高效 edge-aware)+SSIM[0,1], larger better
As illustrated in Figure, we design a novel loss function $\mathcal{L}$, which consists of two main components
\begin{equation}
    \mathcal{L}=\lambda(\alpha\mathcal{L}_{\text{depth}} + \beta\mathcal{L}_{\text{grad}}) \label{overall}
\end{equation}
where $\mathcal{L}_{\text{depth}}$ is a variation of $L_1$ norm of the difference between ground truth and depth estimation, $\mathcal{L}_{\text{grad}}$ stands for the gradient of adjacent pixel blocks, $\lambda$ is a function of SSIM, $\alpha$ and $\beta$ are the hyperparameters.

% \textcolor{red}{This definition seems not correct. What is the difference between this L1 and $L_{depth}$}
$L_1$ norm of the difference between ground truth depth map $y$ and the prediction $\hat{y}$ is the most standard loss function for depth regression problems
\begin{equation}
\mathcal{L}_{1} = \frac{1}{N}\sum_{i=1}^N|y_i-\hat{y}_i|.\label{l1}
\end{equation}
where $N$ stands for the total number of pixel.
However, an issue of directly using $L_1$ norm is that the difference of each pixel, that is, $|y_i-\hat{y}_i|$ for each $i$, has an equal contribution to $\mathcal{L}_{1}$ between distant and nearby pixel~\cite{lee2018single}. For example, the error of 10cm should mean differently for objects at a distance of 1 meter and 10 meters.
Follow~\cite{hu2019revisiting,huynh2020guiding}, we use a logarithmic variation of $\mathcal{L}_{1}$ to alleviate this issue
\begin{equation}
\mathcal{L}_{\text{depth}}=\frac{1}{N}\sum_{i=1}^N F(|y_i-\hat{y}_i|)\label{loss:depth}
\end{equation}
where $F(x)=\ln(x+\theta)$, $\theta$ is a hyperparameter.
There are other methods to deal with this issue, such as using reciprocal of depth~\cite{alhashim2018high} and depth-balanced Euclidean loss~\cite{lee2018single}.

\begin{figure}[ht]
  \centering
  \includegraphics[width=0.9\linewidth]{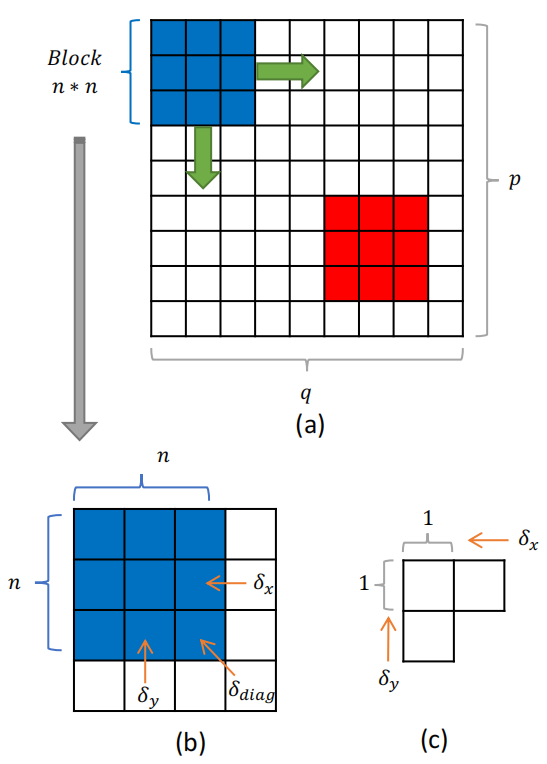}
  \caption{The strategy of calculating $\mathcal{L}_{grad}$. (a) stands for the overall process of gradient calculation for one image: pixel block $b\times b$ slides from left to right, then from top to bottom and the step size is 1. The blue block and red block stands for the starting position and end position respectively. For each step, the calculation of each step is shown as (b). After computing all $\delta_x$, $\delta_y$ and $\delta_{diag}$, we calculate the mean of these three sets of gradient values. (c) is the traditional gradient calculation algorithm, that is, only considering the gradients in x and y directions and unit is fixed to one single pixel.}
  %a表示整体的计算情况，对于不同的channel，都采用这样的方式：块类似滑动窗口，每次步长为1，分别计算三个方向的梯度值之后计算平均。b表示每一步的计算。c表示传统的梯度计算方式，即固定单像素参与计算以及只计算xy两个方向。
  %{\color{blue}{Make sure the figure is in the middle}} }
  \label{fig:loss}
\end{figure}

In addition, in order to track the depth changing between adjacent pixels, we design $\mathcal{L}_{\text{grad}}$
\begin{equation}
\begin{split}
    \mathcal{L}_{\text{grad}} =& \frac{1}{N}\sum_{i=1}^N[ F(\delta_x(y_i)-\delta_x(\hat{y}_i)) + 
    F(\delta_y(y_i)-\delta_y(\hat{y}_i)) \\+&F(\delta_{\text{diag}}(y_i)-\delta_{\text{diag}}(\hat{y}_i))].
   \end{split}
\end{equation}
where $\delta_x$, $\delta_y$ and $\delta_{\text{diag}}$ stands for the mean of the adjacent pixel' gradient of one image in $x$, $y$ and diagonal directions, respectively. 
Specifically, for each direction, the gradient is calculated in the same way, that is, for each pixel $i$, calculate the difference between the value of the next pixel in the current direction and $i$'s value and then the gradient of this direction is the mean of all the differences.
Different from previous work~\cite{huynh2020guiding, alhashim2018high}, we originally incorporate the diagonal components into the gradient calculation. 
Considering the complexity of the shape of real-world objects, our novel $\mathcal{L}_{\text{grad}}$ is able to further penalize small structural errors and improve fine details of depth maps. 
Apparently, this kind of loss is able to effectively reduce grid artifacts and blurry edges. 
However, one more dimension of computing increases the requirements for computing resources, which has a huge impact on resource-constrained areas. 
Therefore, we propose a trade-off by computing the gradient between adjacent pixel blocks, rather than a single pixel, for each pixel blocks $b\times b$, we use the mean of each channel of all pixels in it to represent, where size $b$ is a hyperparameter. 
Figure.~\ref{fig:loss} shows the detail of this strategy. 
It should be pointed out that~\cite{hu2019revisiting,huynh2020guiding} also consider to further improve details, however, their methods rely on the expensive inner product of vectors, which is unaffordable for resource-constrained areas.

Lastly, we add a coefficient $\lambda=1-{\rm SSIM}(y,\hat{y})$ to our overall loss function.
The SSIM is a well-known quality metric used to measure the similarity between two images and is considered to be correlated with the quality perception of the human visual system~
\cite{chen2021attention}.
${\rm SSIM}(x,y)$ is a real number in the unit interval, and the larger its value is, the more similar the two images are. 
Therefore, we use $\lambda$, instead of ${\rm SSIM}(x,y)$, in our loss function.

\subsection{Data Augmentation}
For better generalization in computer vision related tasks, data augmentation is necessary, which is an effective strategy to reduce over-fitting of CNNs~\cite{krizhevsky2012imagenet}.
Vertical flip and horizontal flip are most common strategies of data augmentation. 
Therefore, we execute vertical and horizontal flipping on images at a probability of $\zeta$ and $\eta$ respectively, where $\zeta$ and $\eta$ are hyperparameters. 
%In addition, we also swap the green and red channels on the input images at a probability of $\theta$, where $\theta$ is also a hyperparameter.
As described in~\cite{alhashim2018high}, despite image rotations and distortions are also common data augmentation methods, 
they introduce useless information for the ground-truth depth, such as unnecessary geometric interpretations and invalid data.
Therefore, we do not include these two methods in our approach.

\section{Experiments}
\label{subsec:ex}
We test the effectiveness of our approach separately on different datasets, and compare it with several representative baseline methods, which can represent the STOA. 
Then we provide the ablation study that evaluates the contribution of each component described in Section IV.

\subsection{Datasets}
We evaluate our method on KITTI~\cite{geiger2012we} and NYU-v2~\cite{silberman2012indoor} datasets, which are the most commonly used datasets for monocular depth estimation in computer vision.

\textbf{KITTI} is an outdoor dataset for monocular deep estimation and object detection and tracking based on deep learning, which is captured through a car equipped with 2 high-resolution color cameras, 2 gray-scale cameras, laser scanner and global positioning system (GPS) and contains 93,000 training samples. The original image size is around 1,242 $\times$ 375, and its ground-truth depth maps are sparse with a lot of missing data. Therefore, we execute inpainting method to fill the missing parts~\cite{geiger2012we}.
We use the training/ testing sets split of Eigen et al.~\cite{eigen2014depth}, which is the most standard method for KITTI splitting. 

\textbf{NYU-v2} focuses on the indoor scenes, which contains about 120K frames of RGB-D image pairs captured by a RGB camera and the Microsoft Kinect depth camera to simultaneously collect the RGB and depth information. the original image size is 640 $\times$ 480.
Similar with KITTI, we also execute inpainting method to fill missing depth values.
We follow the official training/ testing split, which uses 249 scenes for training and 215 scenes (654 images) for testing. From the total 120K image-depth pairs, we train our model on a 50K subset as~\cite{alhashim2018high}.

\subsection{Metrics}
\textbf{KITTI. }Following~\cite{eigen2014depth}\cite{ming2021deep}, we adopt both error and accuracy metrics to evaluate the performance of depth estimation methods. The error metrics (\textbf{the smaller the better}) include root mean square error (RMSE), the logarithm root relative error (log.rel), absolute relative error (abs.rel) and square relative error (sq.rel). And the accuracy metrics (acc, \textbf{bigger is better}) include $\delta_1<1.25$, $\delta_2<1.25^2$ and $\delta_3<1.25^3$. These metrics are formulated as:
%\begin{equation}
\begin{align*}
    &\text{RMSE}=\sqrt{\frac{1}{|N|}\sum_{i\in N}||d_i-d_i^*||^2},
    \\&\text{log.rel}=\frac{1}{|N|}\sum_{i\in N} \left|\log_{10}(d_i)-\log_{10}(d_i^*)\right|,
    \\&\text{abs.rel}=\frac{1}{|N|}\sum_{i\in N}\frac{|d_i-d_i^*|}{d_i^*},
    \\&\text{sq.rel}=\frac{1}{|N|}\sum_{i\in N}\frac{||d_i-d_i^*||^2}{d_i^*},
    \\&\text{acc: \% of } d_i \text{ s.t. } \max{\left(\frac{d_i}{d_i^*},\frac{d_i^*}{d_i}\right)}=\delta<\delta_t
\end{align*}
where $d_i$ and $d_i^*$ stand for the predicted depth value and the ground truth value of pixel $i$, respectively, $N$ denotes the total number of pixels with real-depth values, and $\delta_t$ means the threshold for $t\in\{1,2,3\}$.
%\end{equation}

\textbf{NYU-v2. }Following standard practice~\cite{fu2018deep,alhashim2018high}, we use the same metrics with KITTI except sq.rel.
%\subsection{Baselines}

\subsection{Hyperparameters}
In our experiments, we set $\zeta=0.3$ and $\eta=0.3$ for the probabilities of vertical and horizontal flipping in data augmentation.
We use the ADAM optimizer with learning rate 0.0001 and the batch size is set to 8.
For effectively combining $\mathcal{L}_{depth}$ with $\mathcal{L}_{grad}$ as Equation~\ref{overall}, we experimentally set $\alpha=1$ and $\beta=0.8$.
We set $\theta=0.5$ for the logarithmic function $F(\cdot)$ used in Equation~\ref{loss:depth}.
As for the block size $b$, we set it to 2 according to comprehensive experiments.
We set $p=3$ for power average pooling. 

\begin{table*}[!tbp]

\caption{Comparisons of different methods on NYU. In each column we bold the best performing method and underline the second-best.}
\begin{center}
\begin{tabular}{|c|ccc|ccc|}
\hline
Methods           & \multicolumn{1}{c|}{$\delta_1$} & \multicolumn{1}{c|}{$\delta_2$} & \multicolumn{1}{c|}{$\delta_3$} & \multicolumn{1}{c|}{RMSE} & \multicolumn{1}{c|}{log. rel} & \multicolumn{1}{c|}{abs. rel}  \\ \hline
                  & \multicolumn{3}{c|}{(Higher is better)}          & \multicolumn{3}{c|}{(Lower is better)}                                                           \\ \hline
Eigen at al.~\cite{eigen2014depth}               & \multicolumn{1}{c|}{0.769} & \multicolumn{1}{c|}{0.950} & \multicolumn{1}{c|}{0.988} & \multicolumn{1}{c|}{0.641}     & \multicolumn{1}{c|}{-}   &\multicolumn{1}{c|}{0.158}        \\ \hline

Fu et al.~\cite{fu2018deep}               & \multicolumn{1}{c|}{0.828} & \multicolumn{1}{c|}{0.965} & \multicolumn{1}{c|}{\underline{0.992}} & \multicolumn{1}{c|}{0.509}     & \multicolumn{1}{c|}{\underline{0.051}}   &\multicolumn{1}{c|}{\underline{0.115}}        \\ \hline

Alhashim et al.~\cite{alhashim2018high}               & \multicolumn{1}{c|}{\underline{0.846}} & \multicolumn{1}{c|}{\underline{0.974}} & \multicolumn{1}{c|}{0.990} & \multicolumn{1}{c|}{\underline{0.465}}     & \multicolumn{1}{c|}{0.053}   &\multicolumn{1}{c|}{0.123}        \\ \hline

Laina et al.~\cite{laina2016deeper}               & \multicolumn{1}{c|}{0.811} & \multicolumn{1}{c|}{0.953} & \multicolumn{1}{c|}{0.988} & \multicolumn{1}{c|}{0.573}     & \multicolumn{1}{c|}{0.055}   &\multicolumn{1}{c|}{0.127}        \\ \hline

Hao et al.~\cite{hao2018detail}               & \multicolumn{1}{c|}{0.841} & \multicolumn{1}{c|}{0.966} & \multicolumn{1}{c|}{0.991} & \multicolumn{1}{c|}{0.555}     & \multicolumn{1}{c|}{0.053}   &\multicolumn{1}{c|}{0.127}        \\ \hline

MS-CRF et al.~\cite{xu2017multi}               & \multicolumn{1}{c|}{0.811} & \multicolumn{1}{c|}{0.954} & \multicolumn{1}{c|}{0.987} & \multicolumn{1}{c|}{0.586}     & \multicolumn{1}{c|}{0.052}   &\multicolumn{1}{c|}{0.121}        \\ \hline

Ours              & \multicolumn{1}{c|}{\textbf{0.855}} & \multicolumn{1}{c|}{\textbf{0.980}} & \multicolumn{1}{c|}{\textbf{0.994}} & \multicolumn{1}{c|}{\textbf{0.441}}     & \multicolumn{1}{c|}{\textbf{0.047}}   &\multicolumn{1}{c|}{\textbf{0.107}}        \\ \hline
\end{tabular}
\label{NYU}
\end{center}
\end{table*}

\begin{table*}[!tbp]

\caption{Comparisons of different methods on KITTI. In each column we bold the best performing method and underline the second-best.}
\begin{center}
\begin{tabular}{|c|ccc|cccc|}
\hline
Methods           & \multicolumn{1}{c|}{$\delta_1$} & \multicolumn{1}{c|}{$\delta_2$} & \multicolumn{1}{c|}{$\delta_3$} & \multicolumn{1}{c|}{RMSE} & \multicolumn{1}{c|}{log. rel} & \multicolumn{1}{c|}{abs. rel} & sq. rel \\ \hline
                  & \multicolumn{3}{c|}{(Higher is better)}          & \multicolumn{4}{c|}{(Lower is better)}                                                           \\ \hline
Godard et al.~\cite{godard2017unsupervised}     & \multicolumn{1}{c|}{0.861} & \multicolumn{1}{c|}{0.949} &0.976  & \multicolumn{1}{c|}{4.935}     & \multicolumn{1}{c|}{0.206}        & \multicolumn{1}{c|}{0.114}        &0.898        \\ \hline
Eigen at al.~\cite{eigen2014depth}      & \multicolumn{1}{c|}{0.692} & \multicolumn{1}{c|}{0.899} & 0.967 & \multicolumn{1}{c|}{7.156}     & \multicolumn{1}{c|}{0.270}        & \multicolumn{1}{c|}{0.190}        &  1.515      \\ \hline
Kuznietsov et al.~\cite{kuznietsov2017semi} & \multicolumn{1}{c|}{0.862} & \multicolumn{1}{c|}{0.960} &0.986  & \multicolumn{1}{c|}{4.621}     & \multicolumn{1}{c|}{0.189}        & \multicolumn{1}{c|}{0.113}        &0.741        \\ \hline
Alhashim et al.~\cite{alhashim2018high}         & \multicolumn{1}{c|}{0.886} & \multicolumn{1}{c|}{0.965} &0.986  & \multicolumn{1}{c|}{4.170}     & \multicolumn{1}{c|}{0.171}        & \multicolumn{1}{c|}{0.093}        &0.589        \\ \hline
Fu et al.~\cite{fu2018deep}         & \multicolumn{1}{c|}{\underline{0.932}} & \multicolumn{1}{c|}{\underline{0.984}} &\underline{0.994}  & \multicolumn{1}{c|}{\underline{2.727}}     & \multicolumn{1}{c|}{\underline{0.120}}        & \multicolumn{1}{c|}{\underline{0.072}}        &\underline{0.307}        \\ \hline
Ours              & \multicolumn{1}{c|}{\textbf{0.947}} & \multicolumn{1}{c|}{\textbf{0.989}} & \textbf{0.996} & \multicolumn{1}{c|}{\textbf{2.548}}     & \multicolumn{1}{c|}{\textbf{0.113}}        & \multicolumn{1}{c|}{\textbf{0.061}}        &  \textbf{0.297}      \\ \hline
\end{tabular}
\label{KITTI}
\end{center}
\end{table*}
\begin{table*}[!tbp]

\caption{Ablation studies of models without and with different components on KITTI. This shows the importance of all components described in Section~\ref{subsec:pm} in guiding our model.}
\begin{center}
\begin{tabular}{|c|ccc|cccc|}
\hline
           & \multicolumn{1}{c|}{$\delta_1$} & \multicolumn{1}{c|}{$\delta_2$} & \multicolumn{1}{c|}{$\delta_3$} & \multicolumn{1}{c|}{RMSE} & \multicolumn{1}{c|}{log. rel} & \multicolumn{1}{c|}{abs. rel} & sq. rel \\ \hline
                  & \multicolumn{3}{c|}{(Higher is better)}          & \multicolumn{4}{c|}{(Lower is better)}                                                           \\ \hline
Without $\lambda$     & \multicolumn{1}{c|}{0.932} & \multicolumn{1}{c|}{0.977} &0.991  & \multicolumn{1}{c|}{2.831}     & \multicolumn{1}{c|}{0.121}        & \multicolumn{1}{c|}{0.071}        &0.346        \\ \hline

Without diagonal gradient     & \multicolumn{1}{c|}{0.933} & \multicolumn{1}{c|}{0.979} & 0.995 & \multicolumn{1}{c|}{2.769}     & \multicolumn{1}{c|}{0.119}        & \multicolumn{1}{c|}{0.068}        &  0.319      \\ \hline

Without gradient & \multicolumn{1}{c|}{0.917} & \multicolumn{1}{c|}{0.972} &0.988  & \multicolumn{1}{c|}{3.457}     & \multicolumn{1}{c|}{0.179}        & \multicolumn{1}{c|}{0.097}        &0.466        \\ \hline

Without CBAM       & \multicolumn{1}{c|}{0.814} & \multicolumn{1}{c|}{0.957} &0.979  & \multicolumn{1}{c|}{4.266}     & \multicolumn{1}{c|}{0.201}        & \multicolumn{1}{c|}{0.139}        &0.832        \\ \hline

Ours              & \multicolumn{1}{c|}{0.947} & \multicolumn{1}{c|}{0.989} & 0.996 & \multicolumn{1}{c|}{2.548}     & \multicolumn{1}{c|}{0.113}        & \multicolumn{1}{c|}{0.061}        & 0.297       \\ \hline
\end{tabular}
\label{ablation}
\end{center}
\end{table*}
\subsection{Analysis}
\label{subsec:ana}
\textbf{Quantitative comparison.}
In Table~\ref{NYU} and Table~\ref{KITTI}, we compare the proposed algorithm with the recent SOTA algorithms\cite{fu2018deep, alhashim2018high, laina2016deeper,hao2018detail, xu2017multi,godard2017unsupervised} and pioneering work~\cite{eigen2014depth} in monocular depth estimation on NYU-V2 and KITTI dataset quantitatively, which is able to fully demonstrate the effectiveness of our method.

Specifically, compared with~\cite{eigen2014depth}, our approach achieves significant improvement in all metrics on both KITTI and NYU-V2 datasets. Especially for KITTI, the error metrics is almost halved and RMSE is even reduced by 4.508. 
As compared with other methods, the overall performance of our approach is still superior. For NYU-V2 dataset, our approach is first place except $\delta_1$ and abs.rel are second place. As for KITTI, our approach shows the best performance in $\delta_1$, $\delta_3$, RMSE, log.rel and abs.sql. In $\delta_2$ and sql.rel, although our method only won the second place, the difference between the first place is tiny (0.001 in $\delta_2$ and 0.006 in sql.rel).

\begin{figure}[!tbp]
\centerline{\includegraphics[width=0.5\textwidth]{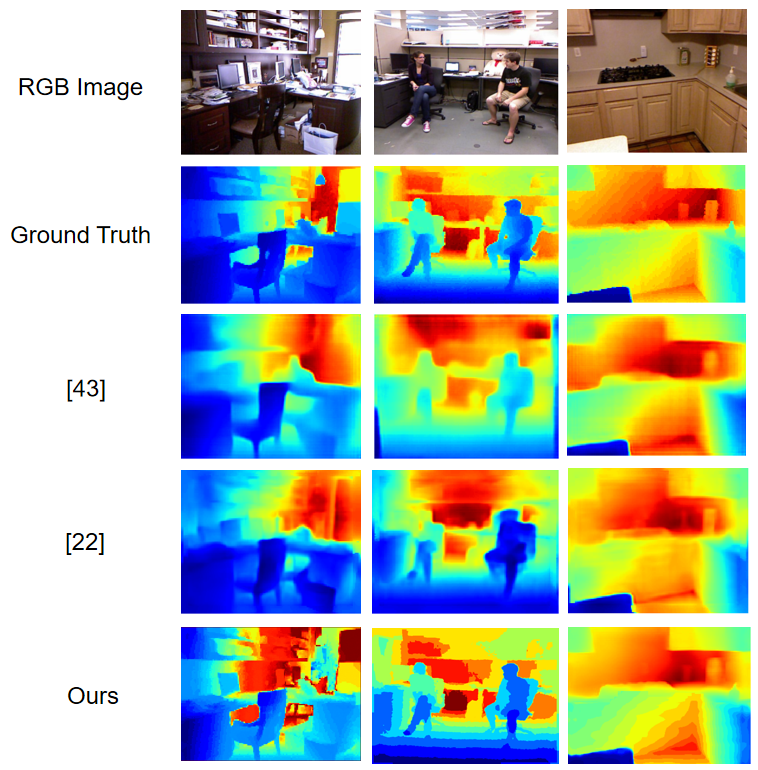}}
\caption{Qualitative comparison of estimated depth maps. We select three representative scenarios (office, classroom with students and kitchen) from the testing set of NYU-V2 for intuitively reflecting the effectiveness of our approach. }
\label{fig:com}
\end{figure}

\textbf{Qualitative comparison.}
Figure~\ref{fig:com} compares depth maps qualitatively. For better visualizations, we transfer original depth maps to color map by calling a toolbox in \textit{matplotlib}.
It is observed that our proposed approach estimates the depth maps reliably and accurately and also reduce grid and blurry edging artifacts in comparison with the other approaches.

\subsection{Ablation Study}
\label{subsec:as}
We present an ablation study to measure the contributions of different components in our approach.
We run experiments on KITTI dataset and the results are shown in Table~\ref{ablation}.

It can be seen that the attention module significantly improves the performance, which demonstrates the effectiveness of applying the attention mechanism to the depth estimation problem. 
In addition, adding gradient components to the loss function also improves performance, where the gradient in diagonal direction further refines the depth estimation results. 
By simply removing term $\lambda=1-{\rm SSIM}$ of our overall loss function~\ref{overall}, we also show that SSIM plays an positive role in this problem.

\section{Conclusion}
\label{subsec:con}
This paper explores the problem of monocular depth estimation based on single image, which is the most extreme but alluring case in vision-related depth estimation. 
We propose an attention-based encoder-decoder network with a novel loss function to effectively address this problem.
Specifically, we insert lightweight attention module CAMB into skip connections between encoder and decoder to find focuses for reducing grid artifacts and blurry edges with least possible overhead. 
In addition, our loss function is designed by combining the difference of depth values, gradient in three directions (x, y and diagonal) and SSIM to further improve fine details of depth maps and penalize small structural errors. 
For speeding up loss computing, we utilize flexible pixel blocks as units of computation instead of single pixel.
Comprehensive experiments on two large-scale datasets show that our approach outperforms several representative baseline methods.

\bibliographystyle{IEEEtran}
\bibliography{reference}

\end{document}